# Cuffless Blood Pressure Prediction from Speech Sentences using Deep Learning Methods

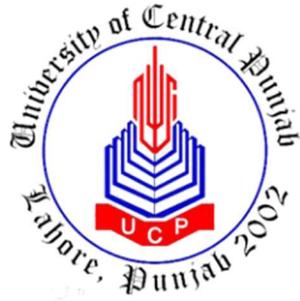

MASTER OF SCIENCE
IN
DATA SCIENCE

Submitted By
Kainat
L1F22MSDS0005

DEPARTMENT OF COMPUTER SCIENCES
FACULTY OF INFORMATION TECHNOLOGY & COMPUTER SCIENCES
UNIVERSITY OF CENTRAL PUNJAB

July 2025

# Cuffless Blood Pressure Prediction from Speech Sentences using Deep Learning Methods

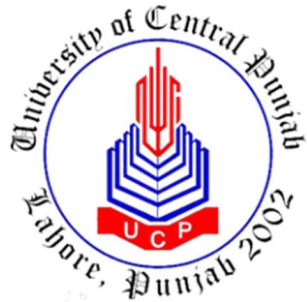

A Thesis submitted in partial fulfillment
of the requirements for the degree of

MASTER OF SCIENCE
IN
DATA SCIENCE

Submitted By
Kainat
L1F22MSDS0005

Supervised By
Dr. Rabia Tehseen

DEPARTMENT OF COMPUTER SCIENCES
FACULTY OF INFORMATION TECHNOLOGY & COMPUTER SCIENCES
UNIVERSITY OF CENTRAL PUNJAB

July 2025

# ABSTRACT

This research presents a novel method for non-invasive arterial blood pressure (ABP) prediction using speech signals, employing a BERT-based regression model. Arterial blood pressure is a vital indicator of cardiovascular health, and accurate monitoring is essential in preventing hypertension-related complications. Traditional cuff-based methods often yield inconsistent results due to factors like white-coat and masked hypertension. Our approach leverages the acoustic characteristics of speech, capturing voice features to establish correlations with blood pressure levels. Utilizing advanced deep learning techniques, we analyze speech signals to extract relevant patterns, enabling real-time monitoring without the discomfort of conventional methods. In our study, we employed a dataset comprising recordings from 95 participants, ensuring diverse representation. The BERT model was fine-tuned on extracted features from speech, leading to impressive performance metrics—achieving a mean absolute error (MAE) of 1.36 mmHg for systolic blood pressure (SBP) and 1.24 mmHg for diastolic blood pressure (DBP), with $R^2$ scores of 0.99 and 0.94, respectively. These results indicate the model's robustness in accurately predicting blood pressure levels. Furthermore, the training and validation loss analysis demonstrates effective learning and minimal overfitting. Our findings suggest that integrating deep learning with speech analysis presents a viable alternative for blood pressure monitoring, paving the way for improved applications in telemedicine and remote health monitoring. By providing a user-friendly and accurate method for blood pressure assessment, this research has significant implications for enhancing patient care and proactive management of cardiovascular health.

# DEDICATION

To my beloved parents,

Your unwavering support, love, and encouragement have been the foundation of my journey. You instilled in me the values of perseverance and dedication, guiding me through every challenge. This thesis is a testament to your sacrifices and belief in my dreams. Thank you for being my greatest inspiration.

# ACKNOWLEDGEMENTS

- All Praises to **GOD Almighty** by whose Grace I was able to complete this research project. Thanks to HIM who blessed me with best at every step of my life.

- I would like to express my deepest gratitude and appreciation to **Prophet Muhammad (S.A.W)** for his profound teachings and exemplary life, which have served as a guiding light and source of inspiration throughout my thesis journey.

- We wish to offer our thanks to **Dr. Amjad Iqbal, Dean, FoIT&CS, UCP** who has been the main icon in our gearing up towards accomplishing something fruitful and helpful for the amelioration of society around us.

- I am deeply grateful to my research supervisor **Dr. Rabia Tehseen, Assistant Professor, FoIT&CS, UCP** for her constant support, and knowledge throughout the thesis process. Her leadership and effort have helped shape the direction and success of this research project. Her valuable comments, suggestions, and generous criticism helped me in writing this Thesis. I am very grateful to her from the core of my heart for her expert guidance and sympathetic attitude. Their mentorship and guidance have been pivotal in my academic and research journey, and I am truly fortunate to have had the opportunity to work under their supervision.

- Next to her, I am grateful to **Dr. Shazia Saqib, Professor, School of informatics and Robotics, Institute of Arts and Culture Studies, Lahore, Pakistan (Ex. Associate Professor, FoIT&CS, UCP)** for her valuable guidance and suggestion in selecting the topic for my thesis. Their insightful input and expertise have been instrumental in shaping the direction and significance of this research project.

- Furthermore, I would like to extend my gratitude to my classmate, **Engr. Muhammad Junaid, Assistant Manager (Tech) at Artificial Intelligence Technology Centre (AITeC), National Centre for Physics (NCP),** for his invaluable guidance and unwavering support throughout this process.

# DECLARATION

I, *Kainat* D/O *Asad Malik*, a student of *"Master of Science in Data Science"*, at **"Faculty of Information Technology & Computer Sciences"**, **University of Central** Punjab (UCP), hereby declare that this thesis titled, *"Cuffless Blood Pressure Prediction from Speech Sentences using Deep Learning Methods"* is my own research work and has not been submitted, published, or printed elsewhere in Pakistan or abroad. Additionally, I will not use this thesis to obtain any degree other than the one stated above. I fully understand that if my statement is found to be incorrect at any stage, including after the award of the degree, the University has the right to revoke my MS/M.Phil. degree.

**Signature of Student:**

**Name of Student:** Kainat

**Registration Number:** L1F22MSDS0005

**Date:**

# PLAGIARISM UNDERTAKING

I solemnly declare that the research work presented in this thesis titled, ***"Cuffless Blood Pressure Prediction from Speech Sentences using Deep Learning Methods"*** is solely my research work, and that the entire thesis has been completed by me, with no significant contribution from any other person or institution. Any small contribution, wherever taken, has been duly acknowledged.

I understand the zero-tolerance policy of the HEC and University of Central Punjab towards plagiarism. Therefore, I as an author of the above titled thesis declare that no portion of my thesis has been plagiarized and that every material used by other sources has been properly acknowledged, cited, and referenced.

I undertake that if I am found guilty of any formal plagiarism in the above titled thesis, even after the award of MS/MPhil. degree, the University reserves the right to revoke my degree, and that HEC and the University have the right to publish my name on the HEC/University website for submitting a plagiarized thesis.

**Signature of Student:**

**Name of Student:**        Kainat

**Registration Number:**    L1F22MSDS0005

**Date:**

# CERTIFICATE OF RESEARCH COMPLETION

It is certified that thesis titled, **"Cuffless Blood Pressure Prediction from Speech Sentences using Deep Learning Methods"**, submitted by **Kainat,** Registration No. **L1F22MSDS0005**, for MS degree at **"Faculty of Information Technology and Computer Sciences"**, **University of Central Punjab (UCP)**, is an original research work and contains satisfactory material to be eligible for evaluation by the Examiner(s) for the award of the above stated degree.

**Dr. Rabia Tehseen**
Assistant Professor
Faculty of IT and CS
University of Central Punjab

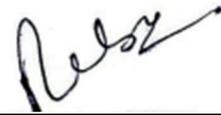

Signature

Date: \_\_\_\_\_\_\_\_\_\_\_\_\_\_\_\_\_\_

# CERTIFICATE OF EXAMINERS

It is certified that the research work contained in this thesis titled **"Cuffless Blood Pressure Prediction from Speech Sentences using Deep Learning Methods"** is up to the mark for the award of **"Master of Science in Data Science".**

**Internal Examiner**

**Signature:**

**Name:** ______________________

**Date:** ______________________

**External Examiner:**

**Signature:**

**Name:** ______________________

**Date:** ______________________

**Dean**
Faculty of IT & CS
University of Central Punjab (UCP)

**Signature:**

**Name:**    Dr. Muhammad Amjad Iqbal

**Date:** ______________________

# TABLE OF CONTENTS

# LIST OF FIGURES

# LIST OF TABLES

# LIST OF ABBREVIATIONS AND ANCRONYM

| | |
|---|---|
| BP | Blood Pressure |
| ABP | Arterial Blood Pressure |
| DBP | Diastolic Blood Pressure |
| SBP | Systolic Blood Pressure |
| CVD | Cardiovascular Disease |
| ECG | Electrocardiogram |
| PPG | Photoplethysmography |
| BERT | Bidirectional Encoder Representations from Transformers |
| MAE | Mean Absolute Error |
| MSE | Mean Squared Error |
| MFCC | Mel-Frequency Cepstral Co-efficients |

# CHAPTER ONE: INTRODUCTION

## 1.1 Arterial Blood Pressure

*The Arterial Blood Pressure (ABP) is an essential indicator of hypertension termed as the pressure of the blood passing through the arteries because of the endeavors of the heart to pump blood.* This pressure is of immense importance in the circulation of blood, delivery of oxygen and nutrients to body parts and organs [1]. This is a fundamental physiological parameter, and it varies under the influence of factors such as cardiac output, vascular stability, overall health of the cardiovascular system and so on. ABP is categorized in systolic blood pressure (SBP) and diastolic blood pressure (DBP) (as shown in **Figure 1.*1***) [1].

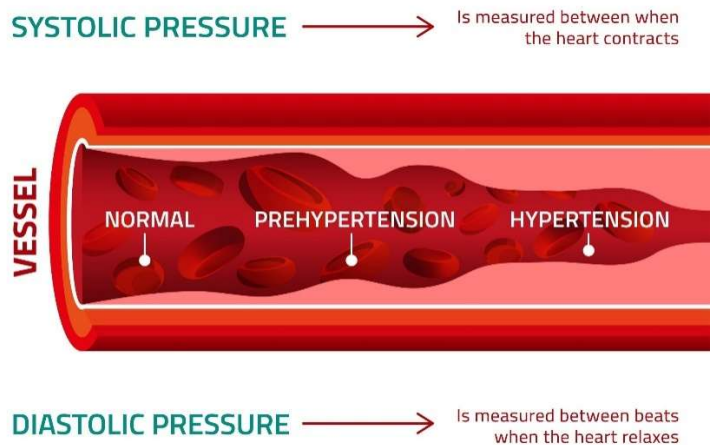

**Figure 1.1: Understanding Blood Pressure Dynamics**

The most notable risk factor of cardiovascular disease (CVD), which constitutes a wide range of conditions aligning to the heart and blood vessels, is hypertension, and hypotension [2]. Monitoring ABP is important to identify and cure hypertension in the early stages and prevent severe health conditions, including heart attacks and strokes [2] [3], [4].

***Systolic Blood Pressure (SBP)*** is an important determinant of cardiovascular health, which determines the utmost amount of pressure within the walls of arteries when the heart is in the contractile stage pumping out the blood into the circulatory system [5]. Elevated systolic pressure is usually accompanied by hypertension, which poses serious health consequences such as heart condition, a stroke, and renal failure. Thus, the SBP needs to be monitored and measured to diagnose and manage hypertension since it gives explanations to the efficiency of the heart and general cardiovascular health. Measurement of SBP is also of particular importance in a clinical setting to assess cardiovascular risks to tailor specific treatments.

Concurrently, ***Diastolic Blood Pressure (DBP)*** represents the lowest point of pressure put on the wall of blood vessels by the heart during the relaxation that occurs in a situation that there are no beatings [5]. Though not as important among the elderly, it is a major indicator of cardiovascular health, particularly among the younger individuals. Variation in DBP may cause underlying health problems such as aerial stiffness, failure of heart or hypotension. The DBP and SBP (as shown in **Figure 1.*2***) [6] play an important role in determining the comprehensive perspective of the cardiovascular status of a person. This serves as a means of early identification of risks and subsequent prevention measure to curb major menaces.

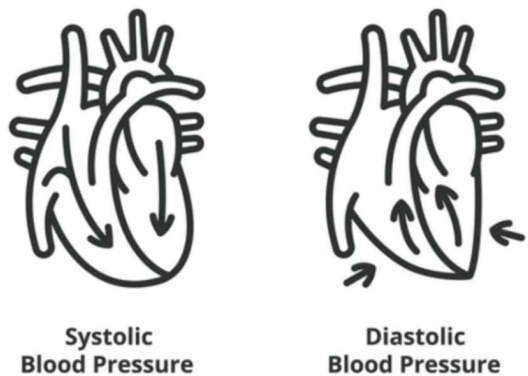

**Figure 1.2: Systolic vs Diastolic Blood Pressure**

Blood pressure readings are typically expressed as a fraction, with 120 mm Hg representing systolic pressure and 80 mm Hg representing diastolic pressure [7]. This notation provides a clear indication of the pressure range in the cardiovascular system during the cardiac cycle [8]. Generally, systolic blood pressure of more than 90 mmHg and a diastolic blood pressure of more than 60 mmHg are considered as the minimum acceptable blood pressure for proper organ function without causing the hypotension systems. However, there are different parameters such as age, health, and medical conditions that can affect these above-mentioned values. Systolic readings below 140 mmHg are normal, while readings between 140-159 mmHg are borderline isolated systolic hypertension, often seen in older adults, and above 160 mmHg is isolated systolic hypertension, a significant risk factor for cardiovascular diseases [9], [10]. Classification of SBP and DBP levels by severity of hypertension is described in **Table 1.1 and Table 1.2** [6], [11], [12], [13].

**Table 1.1: Categorization of Diastolic Blood Pressure (DBP) Levels by Severity of Hypertension**

| Diastolic pressure (mm Hg) | Category |
|---|---|
| <85 | Normal |
| 85–89 | High normal |
| 90–104 | Mild hypertension |
| 105–114 | Moderate hypertension |
| >115 | Severe hypertension |

**Table 1.2: Categorization of Systolic Blood Pressure (DBP) Levels by Severity of Hypertension**

| Systolic pressure (mm Hg) | Category |
|---|---|
| <140 | Normal |
| 140–159 | Borderline isolated systolic hypertension |
| >160 | Isolated systolic hypertension |

Blood pressure can be measured through invasive or non-invasive methods [14], each with its own advantages and disadvantages. Invasive methods are also called inter-arterial BP monitoring, which involve inserting a catheter directly into an artery, usually the radial or femoral artery, to achieve immediate and extremely precise blood pressure measurements [15]. Generally, these types of methods are used in critical care units such as ICU's or during surgical procedures, when an accurate and precise measurement is required (cite: romagnoli2014accuracy). Although these methods offer high accuracy and reliability, but also carry high risks of infections, haemorrhage, and artery damage, making them unsuitable for regular or extended applications. The alternate of these methods are ***non-invasive methods***, are widely used for regular usage such as for home and clinical setups because of their friendly use, safety and convenience. These types of methods do not require breaking the skin or entering the body and are broadly classified into cuff-based and cuffless methods (as shown in **Error! Reference source not found.**) [16], each offering distinct advantages and disadvantages. These methods are particularly important in clinical settings where minimizing discomfort is a priority. Non-invasive methods are further categorized into two main categories: cuff-based and cuffless methods [7].

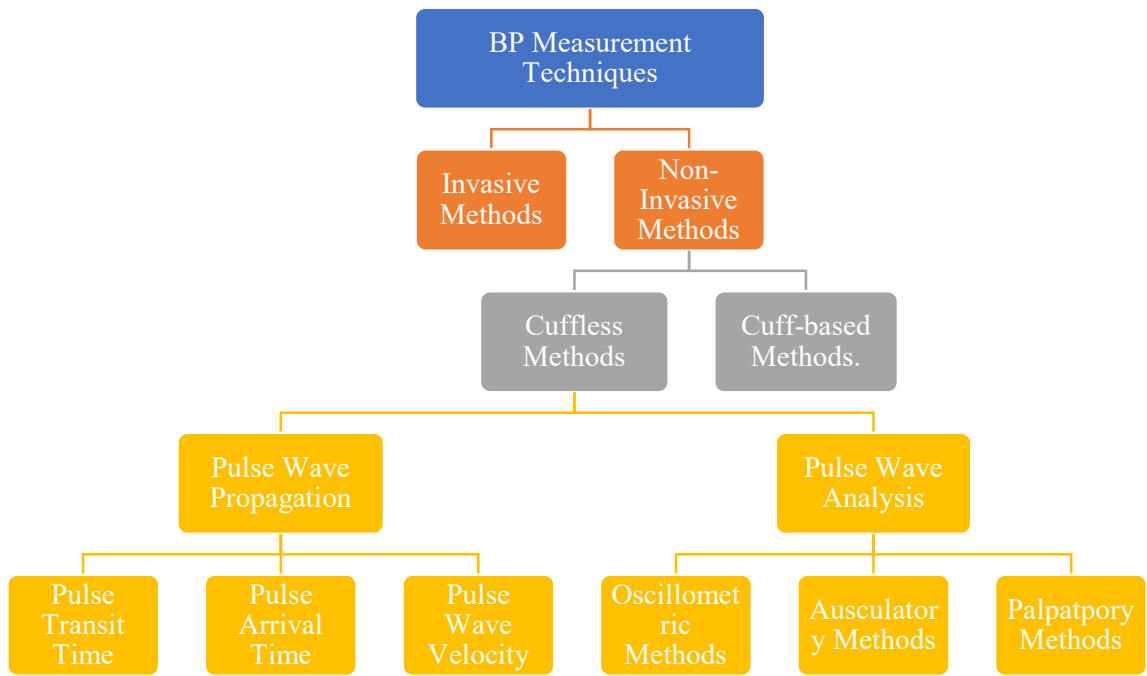

**Figure 1.3: Categorization of BP Measurement Techniques**

This study employs deep learning methodologies, specifically utilizing BERT-based regression techniques, to investigate cuffless blood pressure estimation through speech signals. The research proposes a holistic approach that utilizes advanced natural language processing models to assess acoustic characteristics from voice data for precise blood pressure prediction.

The goal is to provide the non-obtrusive and efficient method of measuring the blood pressure through the use of speech signal analysis and allow constant patient monitoring, as well as their comfort by avoiding the usage of traditional cuff; it also includes making the health monitoring lifelong, by using another part of the body normally involved as a starting point of the process.

## 1.2 Motivation:

The aim of the project is to foretell blood pressure based on speech signal since hypertension and heart diseases are gaining popularity. To improve health outcomes of the population, the under-researched components of the project comprise the inadequacy of traditional blood pressure use methods, the study of the potential use of speech analysis, and the advancements in the field of deep learning technology. Around one-third of adult nationalities all over the world have hypertension that predisposes to heart disease and stroke. The traditional cuff-based devices are uncomfortable and cumbersome and therefore not easy to follow by the individuals with limited mobility or facility access to medical attention to routinely follow the procedure.

There is growing demand in measurement techniques which are non-invasive of simple to monitor the blood pressure and cuffless techniques seem to be a potential solution. Speech signals to predict blood pressure is a novel concept, and it capitalizes on the prevalence of speech in social dialogues which is easily adaptable to fit into the life of individuals. Vocal traits can reflect physiological states that are linked to blood pressure, including heart rate, stress levels, and breathing patterns. By examining these acoustic characteristics, important information about cardiovascular health can be obtained, providing a convenient and non-invasive way to monitor health. Because deep learning technologies, especially BERT models, can accurately predict blood pressure levels across a variety of emotional, physical, and environmental contexts and understand word context, they are essential in improving the viability of speech for BP prediction.

By making health monitoring tools more accessible, the research seeks to enhance both public health objectives and individual health monitoring. Global health outcomes, user compliance, and health monitoring experiences can all be enhanced by combining deep learning technologies with speech analysis. Cuffless blood pressure prediction is still difficult, though, because of problems with accuracy and consistency as well as factors like age, gender, and health. Real-time blood pressure monitoring systems require efficient processing power, particularly in wearable technology. Effective monitoring also requires user acceptance and compliance.

Motivated by the challenges associated with traditional BP measurement techniques, we have introduced a novel method utilizing BERT Regression for cuff-less blood pressure monitoring and assessment using speech signals. This method capitalizes on the advanced capabilities of Bidirectional Encoder Representations from Transformers (BERT), a cutting-edge natural language processing model, to scrutinize and interpret intricate physiological data. Using non-invasive signals, our system is aimed at working on features to increase both the accuracy and reliability of the blood pressure predictions with the help of the regression method. The new system not only addresses the limitations of existing cuff-less methods but also brings in the possibility to monitor it in real-time, a breakthrough in the management of hypertension. By a comprehensive process of validation and testing, we intend to present the effectiveness of this technique to provide accurate blood pressure assessment and thus, would assist in improved patient care and accessibility in blood pressure screening.

## 1.3 Cuff-based BP Prediction:

*Cuff-based methods typically involve an inflatable cuff wrapped around the underarm, which temporarily restricts the blood flow to measure the systolic and diastolic blood pressure.* (as shown in **Error! Reference source not found.**) [16].

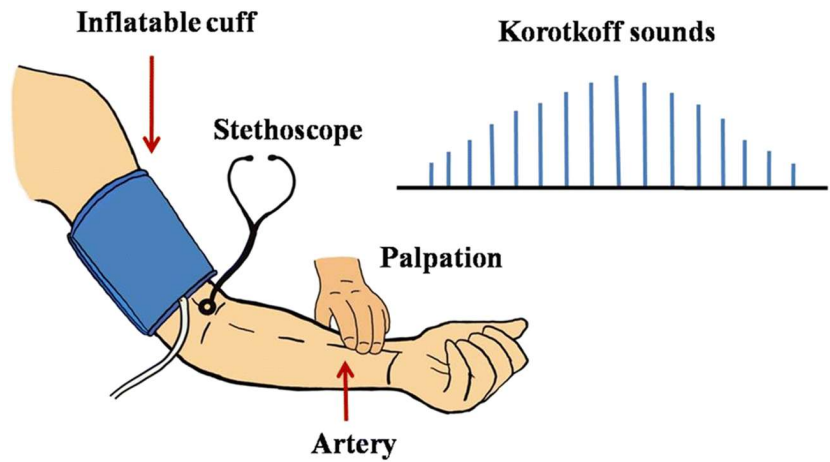

**Figure 1.4: Cuff-based Method for BP Measurement**

These methods involve two primary approaches such as auscultatory and oscillometric methods [17]. Auscultatory method is an accurate way of blood pressure by using sphygmomanometer and stethoscope. In a blood pressure examination, it is done by wrapping an inflatable cuff around the upper arm and measuring the blood pressure taken above the systolic BP enough to temporarily stop blood flow in the brachial artery. Korotkoff sounds are special audio signals indicating turbulence in the blood flow picked up by medical practitioners. The following occurrence of a disappearance is referred to as diastolic blood pressure while the initial detection is referred to as the systolic blood pressure. The auscultatory method is a vital component of clinical blood pressure measurement due to the level of readings of this method, which gives correct and consistent results, although they require experience and expertise. The method, however, should require training and proficiency to identify Korotkoff sounds and also the accuracy of measurement could be affected by factors such as inappropriate cuff positioning, background or even hearing acuity [18].

But in finding oscillation along the arterial wall, when blood flow is restored after artery occlusion, the reliable method by Oscillo metric technique is a sure method in which automated blood pressures (BP) monitors determine the oscillation in the arterial wall. The measurement and interpretation of these oscillations gives an estimation of systolic and diastolic readings of blood pressure. This is a process whereby a pressure cuff is slowly removed allowing blood through an artery after inflating to block the flow. The monitor, then, reads the pulsatile blood flow which causes cuff pressure oscillations. The device can find diastolic and systolic pressures by relying on the amplitude of these oscillations. This way of measure is reliable in medica practice due to accuracy and simplicity and it is applicable in home monitoring [19], [20].

Though cuff-based methods are accurate, those patients with physical limitations or ill conditions might not be comfortable using it. They can be uncomfortable during inflation and deflation especially by those with severe obesity, lymphoedema, arm injuries or with physical limitations among other. These challenges notwithstanding, cuff-based methods remain the gold standard in the measurement of blood pressure because of their correctness and reliability. Another common non-invasive method of monitoring blood pressure is cuff-based techniques that interfere with the flow of blood in the body, by surrounding the upper arm with a cuff. The cuff will be deflated until the blood flow is recovered to measure the pressure at the point it returns to flow thus giving the reading of systolic and diastolic pressures. The method is common and easily applicable in the health facility. Non cuff, the methods that utilize modern technologies, such as photoplethysmography and pulse wave analysis are at the dawn of their development and are rarely used in clinical studies. Even though it brings comfort to a patient and a constant check, further research is necessary to establish that they are effective and correct in the clinical field [21].

Nevertheless, the measurement of blood pressure (BP) on cuff may sometimes fail due to the alleged phenomena such as the masked hypertension [22] and white-coat hypertension [23]. Whereas masked hypertension is a condition in which blood pressure levels are normal at physician practice but higher at other occasions, white-coat hypertension is the case at which the patients exhibit elevated blood pressure levels only at physician practice. They are pre-existing conditions that may be influenced by such factors as cuff size and location and occur when patients manifest extreme blood pressure levels in the presence of a physician only [24]. Cuff techniques are commonly used in most situations, and they are effective only where there is constant blood pressure [10]. Considering these problems with cuff-based measurements, a need to develop an accurate and effective method of arterial blood pressure (ABP) monitoring is critical.

Traditional techniques can be shortchanged by the fact that they are sensitive to outside influences and hence they may often show different values, especially in patients whose blood pressure varies. New technological innovations such as wearables and smart sensors hold great potential because they will generate trustworthy solutions that can make the monitoring process easy and comfortable [25]. Devices would be created to monitor the cardiovascular health to allow early interventions into the cardiovascular health of the patient based on the new monitoring systems that offer the high priority of accuracy, usability, and the potential to collect data in the varying conditions.

## 1.4 Cuff-less BP Prediction:

***Cuff-less methods are a modern, non-invasive, and user-friendly method for monitoring blood pressure, using advanced technologies like PPG, ECG, and speech signal analysis to estimate blood pressure without a cuff.***

***Photoplethysmography (PPG)*** is a non invasive light based technique used to measure changes in volume of the blood in microvascular tissue bed. Sensors, which are often used on wearable devices like smart watch, send out light and read the reflected light, which changes dependent on blood flow. Tracing the pulse wave characteristics properly, PPG is often employed in constant monitoring of the heart rate and can be employed in the estimation of blood pressure [26], [27].

Whereas ***ElectroCardioGraphy (ECG)*** records the electrical activity of the heart. ECG assists in blood pressure estimation by evaluating heartbeat timing and comparing it with other physiological indicators. Some wearable technology combines PPG and ECG to improve accuracy and offer more thorough health insights [27]

A novel technique for predicting blood pressure levels is ***speech signal analysis***, which examines at speech patterns and vocal features [28]. The goal of ongoing research is to develop algorithms that can successfully correlate blood pressure readings with speech features. This method shows a lot of promise for non-invasive health monitoring since it presents a unique way for monitoring blood pressure without the necessity of physical sensors. ***(as shown in Error! Reference source not found.)*** [19]***.***

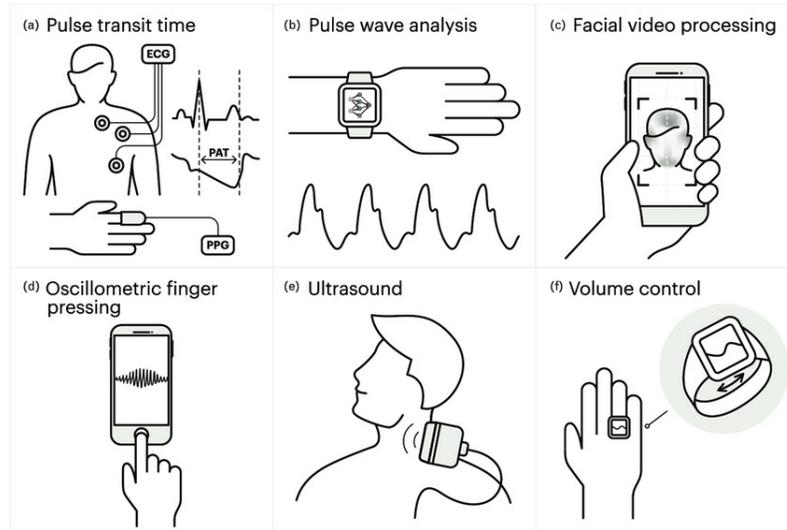

**Figure 1.5: A general picture depicting Cuff-less Method for BP Measurement**

The comfort, convenience, and technological complexity of non-invasive cuff-based and cuff-less techniques differ. While cuff-less approaches present a continuous monitoring solution that is easy to use and constantly evolving, cuff-based techniques are well-established and very accurate.

Speech signal analysis is one of the non-invasive cuffless methods that has a few advantages over more traditional methods like electrocardiography (ECG) and photoplethysmography (PPG) for blood pressure (BP) estimation [29]. It is easy to operate, it does not require body sensors, and it can easily be customized with existing devices such as smart speakers and smartphones. Since analysis of speech is one of the functions that interact with daily lives naturally, there is also the prospect of constant monitoring. The richness of its data is attributed to multifactorial analysis whose results can provide a holistic view of health. Moreover, in contrast with specialised devices, speech analysis can be relatively cheaper since it does not necessitate devices that are costly to acquire [30].

Nevertheless, accuracy and validation can present a challenge to the technology deployed because it is still in the development phase, and noise and pronunciations produced by the environment can challenge the viability of proper measurements.

Even these difficulties do not take away the value of speech signal analysis as a new method to utilise existing technologies and customer behaviour, especially in any environment where ease of use and convenience are important factors [31]. The main applications of cuffless methods of blood pressure prediction are wearable medical technology, telemedicine, mobile health applications, clinical settings, and elder care [32].

The study proposal provides the means of measuring the blood pressure (BP) without a cuff with the aid of speech and deep learning technologies. This voice-non-invasive procedure uses voice acoustics to analyse the speech frequency in search of patterns that have been associated with measures of blood pressure. This will enable real-time continuous blood pressure monitoring, which is more precise, comfortable, and usable on patients. The procedure can be used in several locations (e.g., homes and clinics) where steroid weight machines are not required. Relating to the study, it seems that the combination of deep learning and speech analysis can enhance the ease and accuracy of blood pressure measurement, thus it will contribute to more efficient management of cardiovascular health. The article highlights the importance of real-time predicting blood pressure in early interventions as well as preventive health management. It registers rhythmical speech patterns which are in correlation with the level of blood pressure by the means of BERT (Bidirectional Encoder Representations with Transformers) [33] To make it more accurate in diverse settings in applications like homes and clinics, the methodology retrieves the relevant acoustic characteristics in speech, and trains the models on multiple datasets.

Such an inexpensive method of blood pressure measurement can potentially enhance the accessibility of cardiovascular health surveillance since people will be able to check their blood pressure with greater ease, devoid of cumbersome paraphernalia. Speech analysis together with BERT-based regression can yield to the improvement of hypertension and connected conditions treatment, the consequences of which in the long run will positively influence patient outcomes. The project expects to use BERT-based regression to create an efficient system to predict cuffless blood pressure basing on voice patterns and achieve higher accuracy, effectiveness, and participation of people in their health assessment.

## 1.5    Significance of Cuff-less BP Prediction:

Cuff-less BP prediction plays an important role in our daily life applications.

1. ***Patient Comfort and Experience:*** Omission of inflating a cuff with a cuffless technique also makes it more comfortable for the patient by avoiding inconveniences. Besides this, the new technology helps remove the fearness that is usually attached to the conventional blood pressure measurement procedures, and so, the subject being monitored will feel less terrified. Moreover, the ease of use of this method promotes the high usage of direct and regular health assessments, which further promotes the active involvement of a person in minding his own health [34].

2. ***Increased Accessibility:*** Non-invasive visions are created to be used in remote places with or without specialized equipment, which monitors blood pressure in underserved communities, thus promoting active health care. Health monitoring is more likely to become accessible to people with various backgrounds, particularly via wearables and smartphones [35].

3. ***Continuous Monitoring:*** The techniques facilitate easy collection of data in real time, and this will make the aspect of hypertension management easier. This greatly reduces the incidents of cardiovascular diseases since it helps to diagnose hypertension at its early stages. Additionally, it allows acting on time and changing the course of treatment, which in the end results in an improvement of the quality of patient care and the achievement of improved health [36].

4. ***Improved Accuracy and Reliability:*** Cuffless blood pressure prediction can obtain better readings once improved with the help of artificial intelligence and machine learning. The result of this is better health outcomes, improved management techniques, design of a revolution in how hypertension is monitored and treated [37].

5. ***Cost Effectiveness:*** Cuff-less blood pressure prediction may decrease the costs of healthcare by eliminating the use of expensive equipment and the frequent visits to the doctors. This will enable more people to have access to and affordable care due to the potential of hypertension management using the home or community monitoring process that will relieve patients and healthcare systems of the financial costs required to treat hypertension [38].

6. ***Enhanced Patient Engagement:*** Cuff-less methods improve the level of patient participation in their health management because data can be made to be provided in real-time, frequently measure blood pressure, adherence to the treatment plan, and overall health outcomes are improved [39].

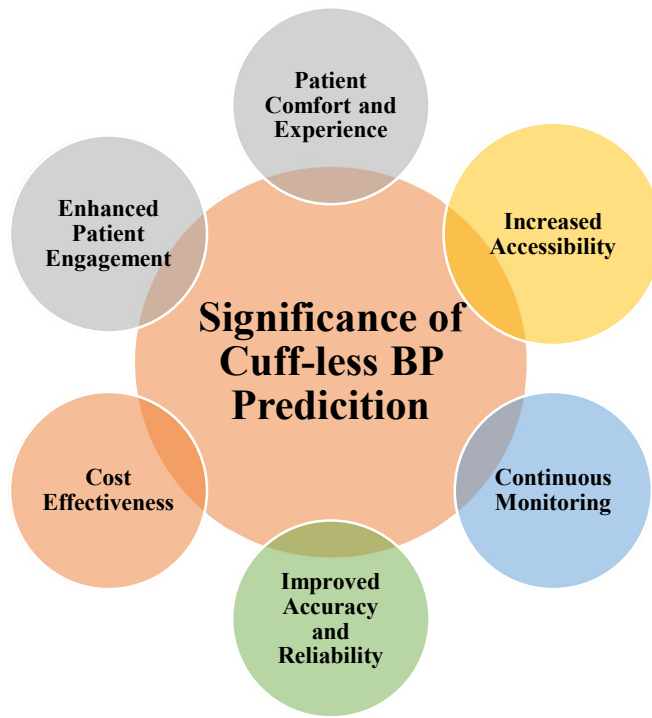

**Figure 1.6: Significance of Crowd Scene Analysis**

## 1.6 Challenges:

Some barriers to cuff-less blood pressure prediction are mainly accuracy, reliability, perceptibility of users and laws and regulations.

1. The fluctuations in reading potential due to external circumstances (e.g. environmental conditions or mobility). Lack of much comparison of traditional techniques of measurement of various populations.

2. It should be easy to combine with current medical systems and apparatus. problems in conformizing sensor technologies across platforms.

3. The patients who have got accustomed to the traditional methods of getting their blood pressure may be adamant.

4. They may lack information or fail to trust the new technology which may slow them down.

5. Regulatory processes to achieve approval and compliance may be difficult to go through.

## 1.7 Problem Statement:

The precise and sustained monitoring of blood pressure (BP) is vital towards the early detection and management of cardiovascular diseases, which is the leading cause of death in the world. Although the existing methods of measurement of BP based on cuffs can be considered clinically reliable, they are neither suitable to be used in real-time nor wearable due to their highly invasive and intermittent nature.

New discoveries have indicated non-invasive examinations with the help of physiological indicators such as ECG, PPG, and speech, and speech is promising because it is natural, contactless, and does not require any special equipment. Nevertheless, the difficulty and manner in which human speech varies, due to linguistic, physiological and emotional factors complications make BP estimation a challenge. Traditional machine learning models of speech in BP prediction frequently use hand developed acoustic features, possibly failing to absorb ground deep semantic and contextual data in natural language expression.

Recent advances of powerful language models such as BERT (Bidirectional Encoder Representations from Transformers) have revolutionized the role of machines in understanding and processing human language; they are capable of highly contextual learning achieving this partially by learning directly on raw texts. Although BERT has proven to be effective in natural language processing tasks, its functionality to interpret the health-related signals, especially the possibility of interpreting the physiological states such as blood pressure and spoken language, has not been quite explored.

This research tries to close such gap by proposing BERT-based model of forecasting systolic and diastolic blood pressure levels using sentences of speech. Using the contextual understanding skills of BERT and combination with the regression models, the study aims at developing an end-to-end (non-invasive and real-time) solution to monitor BP. This form of innovation can usher in new technological changes on passive monitoring of health conditions, making the measurement of BP more available and continuous in both clinical and distant environments.

## 1.8   Aims and Objectives:

The general purpose of our suggested research is to establish dependable non-invasive blood pressure (BP) anticipation estimations given speech declarations. This will be achieved by integrating the recent advancement such as deep learning and signal processing and this will be used to come up with new solutions to check health remotely and preventative healthcare solutions.

Having positive strides in feature engineering to support an array of speech and health-related facial analysis, the primary aim of the present work is the pursuit to analyze and study various acoustic features and combinations, especially in predicting blood pressure. This research would enable higher accuracy and better reliability of cuff-less blood pressure estimation algorithms by evaluation and examination of these many characteristics.

1. The proposed system that aimed to have the capability of accurately predicting blood pressure (BP) using speech sentences, did not require use of conventional cuff-based devices.
2. To put forward an automatic system to have the real-time forecast of blood pressure through deep learning and signal processing ways using speech records.

3. To suggest a method to ensure the reliability of the system through predicting and reducing abnormal blood pressure forecasts in a real-life situation.

This study aims to fill the gap between the spheres of health surveillance and speech processing to present an easy-to-learn and scale device that will enable continuous blood pressure monitoring. The results of the present research can alter the whole concept of non-invasive health monitoring and make it more effective and feasible to a broader range of individuals.

## 1.8 Research Question:

The goals and motivating factors led to the following research questions.

1. How can deep learning models accurately predict blood pressure (BP) from speech sentences without the use of traditional cuff-based sensors?

## 1.9 Novel Contributions:

An overview of our contributions to blood pressure prediction in particular is provided below:

1. A novel method for blood pressure prediction is proposed using voice signals as a non-invasive medium.

Particularly our contributions to blood pressure prediction can be summarized below:

1. A novel approach is proposed for prediction of blood pressure levels by using the speech signals as a non-invasive medium.
2. 2. To capture the complex relationships and patterns in the speech, a transformer-based architecture is proposed using an improved BERT-regression model.
3. A reliable and non-invasive blood pressure prediction method is proposed, which paves the way for possible applications in telemedicine, remote health monitoring, and patient care.

## 1.10 Stakeholders:

- This research work aims to advance the field of non-invasive health monitoring by developing a cuff-less blood pressure (BP) prediction system using speech sentences and deep learning methods. The study focuses on leveraging innovative approaches to accurately estimate blood pressure without the need for traditional cuff-based devices, enhancing convenience and accessibility for users.

- Research questions and methodologies are explored to optimize deep learning models for precise BP prediction from speech data

- The findings can be utilized by healthcare providers and wearable technology developers to improve remote health monitoring solutions and mitigate risks associated with hypertension and cardiovascular diseases.

- The thesis contributes to the field of biomedical engineering and health technology by providing practical guidance and evidence-based insights for developing cuff-less BP prediction systems. By offering innovative methodologies and robust deep learning techniques, the research aims to enhance the accuracy and reliability of BP estimation, ultimately improving public health outcomes and promoting preventive healthcare practices.

- Emergency response agencies can benefit from this research by gaining insights into real-time health monitoring during critical situations, enabling faster and more informed decision-making. The study provides valuable insights into the relationship between speech patterns, physiological signals, and blood pressure, offering a novel approach to non-invasive health assessment.

Overall, this research seeks to bridge the gap between speech analysis and blood pressure monitoring, offering a scalable and user-friendly solution for continuous health tracking. The outcomes of this study have the potential to revolutionize non-invasive health monitoring, making it more accessible and effective for individuals across diverse demographics.

## 1.11 Outline of the Thesis:

Based on the intended research questions, the whole work is divided into seven chapters. The thesis is structured as below:

- **Chapter One: Introduction**
    - Introduces the concept of blood pressure prediction and its significance in healthcare.
    - Explains different methodologies for blood pressure monitoring and prediction.
    - Discusses the motivation behind the research, emphasizing the importance of accurate BP prediction for patient outcomes.
    - Identifies major challenges in the field and outlines the research goals.
    - Concludes with a summary of the research contributions.

- **Chapter Two: Literature Review**
    - Provides a comprehensive review of existing research on blood pressure prediction.
    - Investigates prior studies, theories, models, and methodologies relevant to the topic.
    - Highlights research gaps in current knowledge and emphasizes key findings from previous publications.

- **Chapter Three: Proposed Framework for Cuff-less Blood Pressure Prediction**
    - Details the theoretical background of cuff-less blood pressure monitoring and its relevance.

- o   Outlines concepts such as sensor technology, data collection methods, and machine learning algorithms.
- o   Describes the end-to-end framework, including experimental design and data pre-processing techniques.
- o   Specifies algorithms and techniques utilized to address the research questions.

- **Chapter Four: Implementation and Experimental Results for BP Prediction**
    - o   Discusses implementation details, evaluation metrics, and datasets used for experiments.
    - o   Presents experimental results, including performance evaluations and comparisons with existing methods.
    - o   Interprets findings, draws conclusions, and discusses the implications of the research questions.
    - o   Highlights limitations or challenges encountered during the experimental phase.

- **Chapter Five: Conclusions, Discussion, and Future Work**
    - o   Summarizes the main findings and contributions of the research.
    - o   Discusses potential avenues for future research in blood pressure prediction and monitoring.

# CHAPTER TWO: LITERATURE REVIEW

A brief review of the current approaches along with their shortcomings for cuff-less blood pressure prediction is presented in this chapter. This detailed discussion will highlight the limitations and research gaps to find research direction for future work.

Due to recent breakthroughs in artificial intelligence (AI), deep learning is being more commonly applied in the medical domain that would accurately predict the blood pressure levels using physiological and behavioral data [11]. Examples of its use include prediction of possible occurrence of hypertension, response to treatment, as well as condition typology depending on blood pressure fluctuations. DL-based models examine both EHR and real-time data and enhance the diagnosis and treatment plan, patient care, and facilitate personalized healthcare solutions [12].

Techniques used in blood pressure (BP) measurement can be invasive and non-invasive. These noninvasive techniques are intermittent techniques, which means that BP is measured and continuous techniques through which it is measured continuously. Intermittent techniques employ cuff-based methods which include palpatory, auscultatory, and Oscillo metric; whereas continuous techniques employ incomplete cuff-based methods which are volume clamp and tonometry, and full cuffless methods pulse wave propagation and its analysis [13]. Despite these developments, blood pressure measurement is still a major problem to medical practitioners and researchers.

The new methodology composed of Context aware network (CAN) involving the ECG and PPG Signals was introduced to capture the relationship effectively [14]. Both ECG and PPG signals are non-invasive and can be easily captured using wearable devices. Signals were pre-processed by following different steps, which involve capturing signals across a two-second window, randomizing the amplitude to reduce bias from sensor attachment, and extracting frequency domain information using the Fourier transform. Researchers have developed a Context Aggregation Network (CAN) to predict the link between processed information and arterial blood pressure accurately. The network uses four input channels for training, including amplitude and phase data from ECG and PPG. Proposed model achieves a high degree of accuracy, with reported root mean square errors (RMSEs) of 6 mmHg for diastolic and 7 mmHg for systolic pressures within acceptable limits compared to current literature.

Joung et al. [10] raises the urgency itself to continuous non-invasive blood pressure (BP) monitoring which should be precise and comfortable and accompanied by the lack of solutions available through the already traditional methods. To calculate blood pressure by skin light absorption changes caused by photoplethysmography, a new CNN model, which is referred to PPG2BP-Net was developed. The model is developed in a subject-independent way, and it is right between individuals. The model meets the criterion established by the Association for the Advancement of Medical Instrumentation (AAMI) hence can be used in clinical settings. The high level of accuracy of the model is achieved due to a large set of subjects 4,185. It can be used in the treatment of high blood pressure and cardiovascular disease status besides other fields in medical practice. The findings of the study will have input into the expanding sphere of the non-invasive blood pressure measurement technologies that play an essential role in the comfort and monitoring ease level of the patients involved in the healthcare system.

Suhas et al. [15] introduces a new method for cuffless blood pressure monitoring, addressing the limitations of existing models that rely on handcrafted features from ECG and PPG signals. The authors introduce an end-to-end model based on transformer architecture, demonstrating the potential of deep learning to improve continuous and non-invasive BP measurement. The model, which uses a contrastive loss-based function for robust training, achieves impressive average mean absolute errors of 1.08 mmHg for systolic blood pressure and 0.68 mmHg for diastolic blood pressure. The study also explores the effectiveness of transfer learning when subject-specific data is limited, revealing the model consistently outperforms state-of-the-art methods.

Choi et al. [16] presents a real-time cuff-free device that measures non-invasive blood pressure, with a microphone and ECG electrodes. Its standout feature is pulse transit time (PTT), the system backbone measurement principle, which eliminates the calibration process and makes the system highly comfortable. The system directly measures the blood pulse waves through radial artery, leading to a mean absolute error (MAE) of 2.72 mmHg +- 3.42 mmHg and 2.29 mmHg + -3.53 mmHg on systolic blood pressure (SBP) and diastolic blood pressure (DBP), respectively. The new means may make the monitoring of cardiovascular health more affordable and ongoing, particularly in cases of outpatients.

Another author proposed a study [17], cuffless SBP and DBP estimation by using and learning features on Photoplethysmogram (PPG) waveforms and derivatives with state-of-the-art deep learning recurrent models. An ideal model with a bidirectional recurrent layer, numerous Long Short-Term Memory (LSTM) layers, and attention mechanism returned a spectacular performance on the task of 942 subjects within the MIMIC II database. The model had Mean absolute errors of 4.51 +/- 7.81 mmHg systolic BP and 2.6 +/- 4.41 mmHg diastolic BP and therefore demonstrates the possibility of effective continuous BP monitoring.

Nour et al. [18] has proposed a technique of cuffless blood pressure prediction basing on the photoplethysmography signal and employing learning techniques of machines. They have proven the usefulness of machine learning algorithms in the medical field by taking 24 important features of PPG data and building an accurate model of estimating systolic and diastolic blood pressure.

Zhang et. al., the authors of a study [40], suggested a new approach, which implies the combination of one-dimensional squeeze and excitation (SENet-LSTM) architecture as a metric. Such technique combines photoplethysmography and ECG to enhance feature extraction. The preprocessing improves the feature learning aspect of the model by filtering any noise and artifacts. These findings are consistent with reference to the British Hypertension Society and AAMI guidelines indicating that the readings on categorizing blood pressure levels are high. The analysis reveals that with systolic blood pressure, 94 percent accuracy and for the diastolic blood pressure, 91 percent were achieved, indicating a possibility of further advanced methods of machine learning in monitoring blood pressure non-invasively.

Some new techniques have been suggested to measure blood pressure based on electrocardiograms (ECG), and photoplethysmography (PPG) data to avoid a conventional cuff. These solutions are meant to offer a more convenient and comfortable method in monitoring one blood pressure, especially to people who consider the conventional method of cuff-based measurements to be invasive or inconvenient. Some of the measures which have been used by researchers in investigating the relationship between physiological signals and the level of blood pressure include Mel-Frequency Cepstral Coefficients (MFCC), Pulse Wave Velocity (PWV), and Pulse Transit Time (PTT).

Despite this potential, these cuffless strategies, however, possess a great challenge; they are usually requiring the application of sensors requiring them to be attached to the body. This device often involves wires, which are bulky and may prove discomforting to the people. These physical links might compromise the user experience such that people will not be keen to use these mechanisms to conduct routine monitoring. Such discomfort can restrict the application of these technologies in the normal context, especially outside clinical care. Moreover, PWV and PTT can be obtained through physiological signals, but the issue is that these two parameters cannot be precisely determined because of personal peculiarities which produce discrepancies in the blood pressure records. Single calibrations are often necessary to be sure. These advanced techniques, due to the sensitivity to personal characteristics and states, might not be as practical as being implemented at large scale in real-life.

## 2.1  Summary:

This chapter reviews BP prediction literature, focusing on cuff-less and cuff-based methods. It highlights both traditional and deep learning-based approaches for these tasks. Most of the research in BP prediction focuses on cuff-based methods, which rely on physical devices to measure BP directly. However, these methods can be intrusive and inconvenient for continuous monitoring. Traditional BP prediction methods face challenges such as variability in physiological signals, noise in data collection, and individual differences in speech patterns, making it difficult to achieve accurate and robust predictions. Therefore, ensuring high-quality, consistent, and reliable annotations is essential for building effective BP prediction models. To overcome the limitations of traditional cuffless blood pressure measurement approaches, particularly those that rely on physically connected sensors, our research presents a new approach utilizing voice signals for blood pressure calculation. This research presents an automatic framework for predicting blood pressure levels, specifically focusing on systolic and diastolic measurements, using advanced deep learning techniques, particularly the BERT model.

Our proposed method leverages the acoustic characteristics of speech, which can be captured noninvasively and without the discomfort of traditional sensors. Vocal characteristics such as pitch, tone, and speech patterns can provide important information about cardiovascular health. The study describes the suggested model's special features, including pre-processing and feature extraction methods that successfully differentiate between systolic and diastolic data, hence improving predicting accuracy.

This method not only improves user comfort by eliminating the need for bulky cables and attachments, but it also streamlines the measuring process, making it more suitable for routine monitoring in non-clinical contexts. Our method aims to reduce the need for individual calibration by minimizing calculations like Pulse Wave Velocity (PWV) and Pulse Transit Time (PTT), offering a more practical and accessible alternative to current blood pressure monitoring methods, thereby transforming blood pressure monitoring. This novel technique has the potential to revolutionize blood pressure monitoring by providing a user-friendly, accurate, and convenient alternative to existing methods. The novel approach offers a convenient, accurate, and user-friendly substitute for existing techniques, which could completely transform blood pressure monitoring.

**Table 2.1: Literature Review Summary with Identified Limitations**

| Author(s) | Year | Method / Input | Model / Technique | Dataset / Subjects | Performance | Limitations |
|---|---|---|---|---|---|---|
| Suhas et al. [14] | 2024 | ECG & PPG | Transformer with contrastive loss | Not specified | MAE: 1.08 (SBP), 0.68 (DBP) | High computational complexity; lacks real-world deployment validation |
| Choi et al. [41] | 2023 | ECG & Microphone | PTT-based real-time system | Not specified | MAE: 2.72 ±3.42 (SBP), 2.29 ± 3.53 (DBP) | Susceptible to ambient noise; limited testing in dynamic or ambulatory environments |
| Khan [17] | 2005 | PPG & derivatives | Bi-LSTM + Attention | MIMIC-II, 942 subjects | MAE: 4.51 ± 7.81 (SBP), 2.6 ± 4.41 (DBP) | High error variance; not tested in real-time or on diverse populations |
| Zhang et al. [40] | 2024 | PPG + ECG | SENet-LSTM | Not specified | Accuracy: 94% (SBP), 91% (DBP) | Model complexity may hinder deployment on wearable devices; real-time testing not explored |
| Njoku et al. [42] | 2022 | ECG & PPG | Context Aggregation Network (CAN) | Not specified | RMSE: 7 (SBP), 6 (DBP) | Short signal window may miss long-term BP trends; lacks population diversity |
| Joung et al. [43] | 2023 | PPG | CNN (PPG2BP-Net) | 4,185 subjects | AAMI-compliant | No detailed breakdown of model's performance under noisy or motion-heavy conditions |
| General Studies | — | Speech, ECG, PPG | MFCC, PWV, PTT | Various | Varies | Sensor attachment still required; calibration often needed; discomfort limits everyday use |

# CHAPTER THREE: METHODOLOGY

The primary focus of this chapter is to showcase the innovative framework for cuff-less blood pressure (BP) prediction, leveraging a range of deep learning techniques, with a specific emphasis on BERT.

## 3.1 Theoretical Background

### 3.1.1 Artificial Intelligence

Artificial Intelligence (AI) refers to a branch of computer science that is focused on creating systems that can assign duties that are generally hard to undertake by humans. The set of functions employed by these activities is very multifaceted, and it includes the ability to learn, reason, solve problems, perceive, and understand language [19]. The use of AI is generally divided into narrow AI, which has a set purpose on voice assistants and image recognition, and general AI, which attempts to mimic more wholesome human intellectual capabilities. This taxonomy is used to present the numerous applications and prospects of AI in a variety of fields that demonstrate the brilliant impact of modern technology [44].

In the sphere of AI, critical subspaces appeared, such as machine learning (ML) and deep learning (DL). Machine learning equips systems with an ability to learn using the data and improve on their performance through time without direct coding. This flexibility and capability to revise the algorithms with the emergence of new data has turned ML into one of the foundations of AI applications [45]. A subset of ML is deep learning, which modifies neural networks with many layers, or deep in short, to capture complex patterns in the data. These new developments in terms of neural network architecture have enhanced the performance of AI systems to a great degree and can deal with a more sophisticated range of tasks [46].

Regarding healthcare, AI has transformed how data is analyzed, interpreted, and applied to clinically make decisions. AIs can now process large quantities of patient data through Electronic Health Records (EHRs), images and physiological signals. This ability assists health care providers in diagnosis, planning of treatment and surveillance of diseases. As an example, AI algorithms can identify trends in patient data that can indicate the development of conditions, making it easier to intervene early and develop a personal treatment plan [47]. One of the new ways AI is applied to healthcare is prediction of blood pressure (BP) based on non-invasive data sources such as voice and speech. The new technology uses speech analysis to have an easy time monitoring cardiovascular health. Studies have shown that minute speech aspects are able to associate with physiological disorders like hypertension. Healthcare professionals can implement a new way of painless and constant control of their health by learning these correlations with help of AI models [48], [49].

In this approach, it is not only the care of patients that is improved, but the proactive management of cardiovascular health can also be encouraged thus preventing complications that relate to hypertension. Implementation of AI in the field of healthcare is a breakthrough in the provisions of medical attention, which can boost performance improvements in the medical care of people and medical processes. The further development of AI solutions highlights the growing number of research and usage opportunities that will one day make healthcare more personalized and effective.

### 3.1.2 Natural Language Processing

Natural Language Processing (NLP) is an acidic research area of artificial intelligence that explores the sophisticated interaction(s) between computers and the human language. It has a wide range of methodologies that endow the machines with the ability to recognize, understand and thus produce human language in a meaningful context. NLP enables computers to process literally volumes of text and enables them to do so using a combination of computation linguistics, machine learning and deep learning, making it possible to use in a variety of fields [50].

One of the most important uses of NLP is in text analysis where it is actually used to discover insights in unstructured data. Such a feature is especially important in the spheres such as healthcare, customer service, and finance. With NLP algorithms, clinical notes and electronic health records are reviewed so that the professionals can have important information to explicitly make a decision regarding the patient. In the aspect of providing customer services, chatbot conversations powered by NLP can easily learn and answer customer questions and eventually provide more satisfied customer experience as well as an efficient operation.

Sentiment analysis, or the ability to detect the emotional tones of the text, is another important aspect of the NLP. The method is widely used in social media tracking, brand management, and market-researching because an organization is able to measure what the people are saying and make the necessary reprisal. By cross-referencing customer reviews and feedback, businesses will have very good information on consumer needs and white spots where it can improve their products and services.

In addition, advances in NLP are also driving voices and speech recognition. Voice mail systems, transcription, and virtual assistants greatly depend on NLP to process speech and convert it to text allowing a smooth interaction between humans and the computer. The scope of NLP that may transform the industry and enhance communication will continue to increase, as the application of NLP moves to a new level that will incorporate more intuitive and efficient solutions filling the gap between human language and machine understanding.

**3.1.2.1 Applications for NLP in healthcare:**

- **Clinical Documentation:** NLP enables clinical documentation procedures to be carried out through automation of extraction of relevant data in unstructured clinical notes. This saves the healthcare providers unnecessary administrative duties thus enabling them to have more time to attend to the patients. As an example, NLP systems may extract essential patient data, i.e., symptoms, diagnoses, and treatment plans in the notes left by physicians.

- **Analysis of Electronic Health Records (EHR):** NLP has the potential to mine large volumes of data in EHRs, giving information on the result and effectiveness of treatments on patients. NLP algorithms can establish the tendencies and trends in textual data which can be used to deliver better clinical decisions and community health campaigns.

- **Patient interaction and support:** Chatbots and virtual assistants are fed with NLP ensuring patient interaction in terms of instant resolutions to the general queries. Patients could have better access to information through these tools as they are able to guide them through administrative processes, question medications, and even symptom checkers.

- **Clinical Decision Support Systems (CDSS):** NLP is used as part of CDSS, AI can collect the relevant information in the medical literature and patient records. This allows healthcare professionals to make favorable decisions relying on the newest research and personal data on each patient, consequently enhancing the level of care.

Sentiment Analysis: NLP can be used to detect sentiments on patient reviews, patient experience and reviews, and even postings on social media would offer insight as to sentiments about the healthcare service. This assists organizations to know how satisfied their patients are and what they should improve

### 3.1.2.2 Applications of NLP in Speech Signal

- **Speech Recognition:** NLP plays a major role in the speech recognition technology that is used to transcribe speech into text. This application comes in handy especially in the medical field where practitioners could record notes and reports using the application making documentation more effective and reducing error rates.

- **Voice Biomarkers:** The recent developments in NLP allow studying the characteristics of the voice to make a diagnosis of certain health issues. An example can be given when a certain quality of speech is correlated with such conditions as hypertension, depression, or brain disorder. Through these features, healthcare providers can understand the condition of the patient.

- **Telemedicine:** NLP can be used in telehealth consultations to increase communication between patients and medical workers. It allows conversational transcription that helps keep records and analyze them effectively. Also, due to NLP, one can detect an emotion or a sentiment that can be communicated during consultations, and this can be used in diagnosis and developing a treatment plan.

- **Patient Monitoring:** NLP can also predict speech patterns on chronic conditions. As an example, any alterations in the voice of a patient can signal exacerbation of respiratory disease or other medical conditions, and thus the intervention can take place in time.

In conclusion, we can say that with Natural Language Processing, the industry is seeing a paradigm shift in terms of ease of communication among the team members, clinical decision making, and coming up with 21st century solutions to patient care. Its uses in speech analysis also expand its application thus providing new opportunities of monitoring health circumstances and enhancing patient-provided interactions. The future of NLP technology in healthcare is bright, and as such its powerful role in enhancing the efficacy of healthcare and the streamlining of processes will be even more in the future as a door opens to even more personalized and effective healthcare strategies.

### 3.1.3 Speech Signal Analysis

Speech signal processing, analysis, and interpretation of all aspects of audio signals that occur at the sound output of a force-vibrating system involved in the production of human speech. This critical process plays a major role in ensuring that machines can understand the language that is spoken and hence a critical component in many usages whose main ones include voice-controlled systems, speech recognition, speaker identification, and health monitoring. As signal speech is necessarily analog, they cannot be subject to analytical computations without the process of digitization and preprocessing them, which allow codes to analyze audio data and get the most relevant patterns and characteristics of the speech.

Critical steps in speech signal processing involve the most important steps, which include pre-emphasis, framing, windowing, Fourier analysis and attack extraction. Pre-emphasis aims at increasing the high frequency parts of speech signal to balance the spectral parameters of the signal. Framing follows it up to divide the continuous speech signal into smaller periods or frames to be analyzed over a span of time. Windowing is a procedure that applies mathematical functions to each frame, limiting the signal disturbances at edges and reducing the limits of precision in the relevant (subsequent) analysis.

***Mel-Frequency Cepstral Coefficients (MFCCs)*** rank among the most common features to use in speech processing. MFCCs describe the perceptual and spectral features of speech with a presentation of the signal in a form that is closely suggestive of human hearing by focusing on frequencies that are important to man. Also, other important characteristics that can be extracted out of the signals speech include ***pitch, formants, energy, zero-crossing rate, jitter and shimmer*** that provide the information into various aspects of the speech signal and in making some inferences into physical and emotional conditions.

The particular interest is in speech characteristics which are established as the not-so-invasive indicators of a few states or rather parameters in the sphere of biomedical signal processing. As an example, minor changes in speech patterns have been seen to be correlated with such health conditions as hypertension, depression, and neurological conditions. Examination of these speech characteristics can allow the creation of models used to predict health outcomes, including the blood pressure level, and open the door to new forms of continuous health monitoring. This does not only improve understanding of the connection between speech and health but also expands the possibilities of using speech signal processing in telemedicine and the management of remote patients, improving access to healthcare and its effectiveness. Following the advances in technology, the uses of speech signal processing in healthcare and everyday lives can be far-ranging and are likely to provide significant improvements in communication between humans and machines and health surveillance.

### 3.1.4  BERT-Base

BERT, the abbreviation that stands for Bidirectional Encoder Representations from Transformers, is one of the most recent language models that were proposed by Devlin et al. in 2018. It works on Transformer architecture but only on the encoder part of it. In contrast to the earlier models that managed the text as unidirectional text, BERT understands the entire sequence on a bi-directional basis, and thus, it can understand the context of words in the light of all the words surrounding it in a sentence.

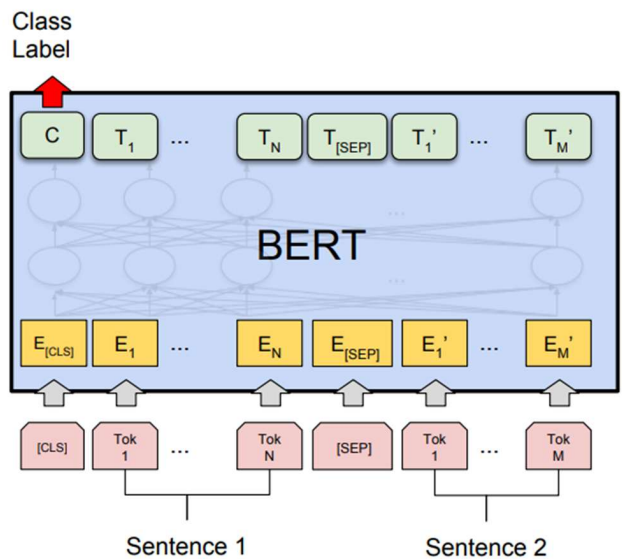

**Figure 3.1: General Architecture of BERT Model**

The BERT-Base has the 12 Transformer encoder blocks, or layers, that possess 768 hidden parts and 12 self-attention conciliations, which makes an aggregate size of 110 million parameters. BERT is trained using large amounts of data such as Book Corpus and Wikipedia through unsupervised training activities:

- Masked language Modeling (MLM): Masked words randomly in a sentence and predicts them using its surrounding context.

- Next Sentence Prediction (NSP): Decides whether consecutive sentences occur in natural text or not.

BERT has shown good performance by being fine-tuned on back-of-the-envelope Natural Language Processing (NLP) tasks with a simple output layer, including classification, question parsing, named entity recognition, etc., after pretraining. Contextual embeddings can be found using BERT on transcripts (created through Automatic Speech Recognition) used in applications that require speech. These embeddings are useful in the next task, such as emotion detecting or estimating physiologic parameters, such as predicting blood pressure.

## 3.2 Materials:

### 3.2.1 Dataset

A publicly available data set [51] is used for training and testing purposes with the underlying characteristics defined in table 1 and 2.

**Table 3.1: Description of Participants taken part in Recording Activity**

| Characteristics | Details |
| --- | --- |
| No. of participants | 95 |
| Age (Min, Max) | 20,70 |
| Inclusion Criteria | Persons with No history of neurological or neuropsychiatric disease; Moderate exercise |
| Exclusion Criteria | Persons with Pregnancy; Drug addiction; Alcohol or caffeine intake within 1 hour prior |
| Recording Details | Stereo recording at 48 kHz with 16-bit resolution |
| Ambient Noise level | 45-65 dB |
| File Format | .wav format |

**Table 3.2: Dataset Description**

| Characteristics | Females | Males |
| --- | --- | --- |
| No. of Subjects | 45 | 50 |
| No. of Speech Recordings | 45 | 50 |
| Heart Rate (Bpm) max, min, average | 111, 46.0, 75.26 | 128, 58.0, 71.25 |
| **Systolic BP (SBP) max, min, average, STD** | 153, 91, 114.28, 15.74 | 153, 86, 119.7, 13.73 |
| **Diastolic BP (DBP) max, min, average, STD** | 98, 35, 77.42, 13.30 | 91, 48, 79.88, 17.01 |
| **Age max, min, average, STD** | 48, 20, 29.9, 9.14 | 46, 20, 29.42, 6.37 |
| **Health Conditions** | | All Healthy |

### 3.2.2 Speech Recordings

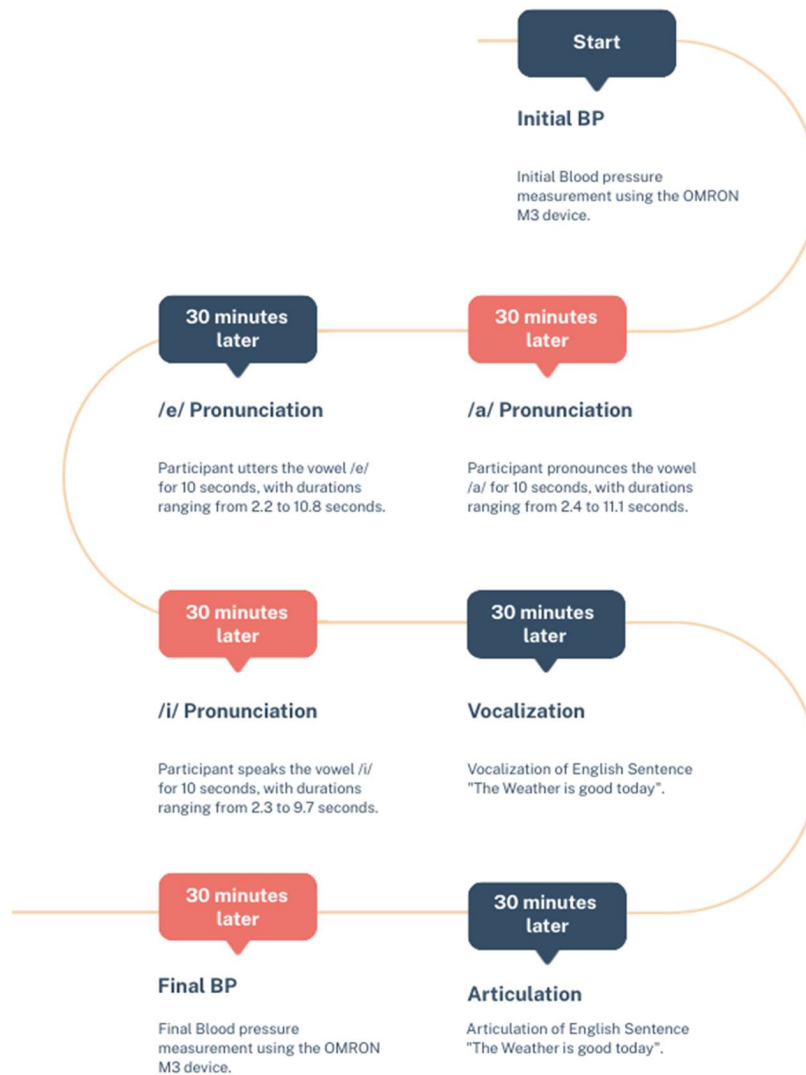

**Figure 3.2: The speech recordings involved a series of vocal exercises and blood pressure measurements over a 3-hour period to explore the relationship between vocalization and blood pressure changes.**

To examine the relationship between speech sentences and blood pressure, different vocal exercises and blood pressure measurements were performed over the duration of 3-hours. The complete process of speech recordings is shown in figure 01, highlighting the systematic approach and attention to detail in data collection. The whole process is divided in 7 different steps, starting from initial BP measurement and ending at final BP measurement.

Omron M3 device was used for the measurement of both readings of the BP measurement. After taking the initial BP measurement, a delay of 30 minutes was taken for participant preparation. Next step involves the pronunciation of vowels "a", "e", and "i" with different durations ranging from 2 to 10 seconds. There was also delay of 30 minutes between each step in order to minimize the external influences and ensure consistency. The next two steps involve the vocalization and articulation of English sentence "The weather is good today", with a similar 30-minute silent interval following this activity. The final step involves the recording of final BP measurement after the interval of 30 minutes.

All of these recordings were saved in .wav format and made in stereo mode at frequency rate of 48 kHz with 16 bit resolution. Noise level was ranging from 45-65 dBs.

## 3.3 End-to-end Proposed Framework:

### 3.3.1 Data Pre-processing

A data set is defined with the following attributes (as shown in the table) based on the initial and final BP measurement measurements. Hypertension category is classified into two categories based on both SBP and DBP threshold values. The threshold values of SBP is assigned as 115mmHg, calculated on the basis of average values initial and final SBP measurements. Similarly, the threshold value of DBP is assigned as 72mmHg calculated on the basis average of initial and final DBP measurements. Participant is marked as hypertensive, if the value goes above SBP/DBP threshold and vice versa. The results are stored in a new feature vector vect_2(), while the participant's ID and names are stored in a separate vector vect_1().

**3.3.2 Audio Processing**

All of the recorded videos undergo normalization process to ensure the consistent amplitude. After the normalization process, all recordings will be the fed to vowel detection and extraction process. This would be useful for estimating blood pressure. Each audio signal will be segmented into smaller parts. Total 2400 numbers of small segments, each of 50mSec generated for detailed feature extraction. Each small segment is then applied to Fast Fourier Transform for extraction of frequency components. These frequency components are then used to identify the spectral peaks, which are crucial for vowel detection. A gaussian windows is also applied for further smoothing and improving the reliability of amplitude estimation.

**3.3.3 Feature Extraction**

In this phase, all the pre-processed audio signals subjected to extract different features responsible for understanding the underlying speech dynamics, such as pitch, spectral and temporal features. These extracted features consist of a range of attributes includes as follows:

3.3.1 **Spectral Features:** These features encompass frequency aspect of speech and they are composed of various key metrics that include Mel-frequency Cepstral co-efficient (MFCCs), spectral centroids, spectral bandwidth and spectral flatness.

3.3.2 **Temporal Features:** Temporal features are important in the analysis of dynamic characteristics of audio signals, especially speech. Important parameters are the rate of zero-crossings (ZCR), energy amplitude and time frames.

3.3.3 **Pitch:** The feature symbolizes the frequency perceived in the audio and it is significant in determining the intonation structures in the speech. Pitch extraction assists in the separation of one vowel sound to another since pitch differences may denote various moods or expressions.

The following features are extracted from the audio signals:

- **Mel-Frequency Cepstral Coefficients (MFCC):** It calculated the spectral analysis of audios taking in the initial 12 MFCC (1-12).
- **Skewness:** It distinguished the tonal and non-tonal characteristics of audios and it gives bias regarding the unevenness of amplitude of distributions.
- **Kurtosis:** The characteristic shows extreme signals and peaks in audio data emphasized by reference to the amplitude distribution as to how tailed it is.
- **PolyArea:** It is also an essential value that should be used to accentuate the energy distribution since it measures the area under the signal curve.
- **Max(x,y) and Min(x,y):** These descriptors give the maximum and minimum amplitude values on the segment, and they give information on the range of a signal.

When these features are obtained, they are systematically joined with concomitant blood pressure data of vect_2(). This connection is very important in forming correlations between speech characteristics and physiological responses. Because of analyzing the relationships between the samples of the audio excerpts and blood pressure parameters we will be able to understand in a more accurate way the possible consequences of the speech patterns in a physiological aspect. This multi-faceted process is beneficial to elicit understanding of how particular characteristics of vocal quality can signify or impact the physiological conditions, which can be of use to the health-related areas where voice analysis is involved.

### 3.3.4 Feature Selection

It is a salient feature of any deep learning algorithm since the filtering out of redundant information can make it less processor-intensive and more accurate as models overall. The reduction of system complexity that is obtained by transformation of the higher-dimensional features into lower dimensional features has the effect of also helping to strengthen the performance of the model. [21], [22]. Feature selection techniques can be classified into three categories including supervised, semi-supervised and unsupervised [23], [2]. Methods like Random Forest, Lasso regularization, forward sequential selection, and ReliefF are used to eliminate irrelevant features in acoustic signal processing \cite{rong2009acoustic}. However, ReliefF, often used in acoustic data classification, outperforms other approaches, achieving higher accuracies [24], [25], [26]. We used ReliefF to reduce the feature vector by eliminating the irrelevant and negative features. The feature selection process was optimized using 10-fold cross-validation was used to optimize the overall process of feature selection.

According to the analysis, features like Max(x,y) and Min(x,y) are less significant than Mel Frequency Cepstral Coefficients (MFCCs), skewness, kurtosis, and polygonal area weights. Features that improve model accuracy are indicated by positive weights, and features that have a negative effect on accuracy are indicated by negative weights. For the ML algorithm, Max(x,y) and Min(x,y) carry extraneous and harmful information. As indicated by their larger weights, MFCCs have the greatest impact on the model's performance.

### 3.3.5 Proposed Method:

In our study, we proposed a comparative framework for cuffless BP prediction using English speech sentences based on BERT Regression, utilizing its unique strengths to accurately predict BP from audio features.

It uses transformer-based models to understand speech sentence context and semantics. It involves transcribed speech sentences into text format, tokenized using BERT's tokenizer, and generating contextual embeddings for input tokens. A regression head is added to the model to predict BP values, fine-tuned on the dataset to learn the relationship between linguistic features and BP levels. Like the CNN approach, the BERT model is trained using a labeled dataset and evaluated through cross-validation techniques. The fine-tuned BERT model predicts BP values from new speech sentences, leveraging its understanding of language context.

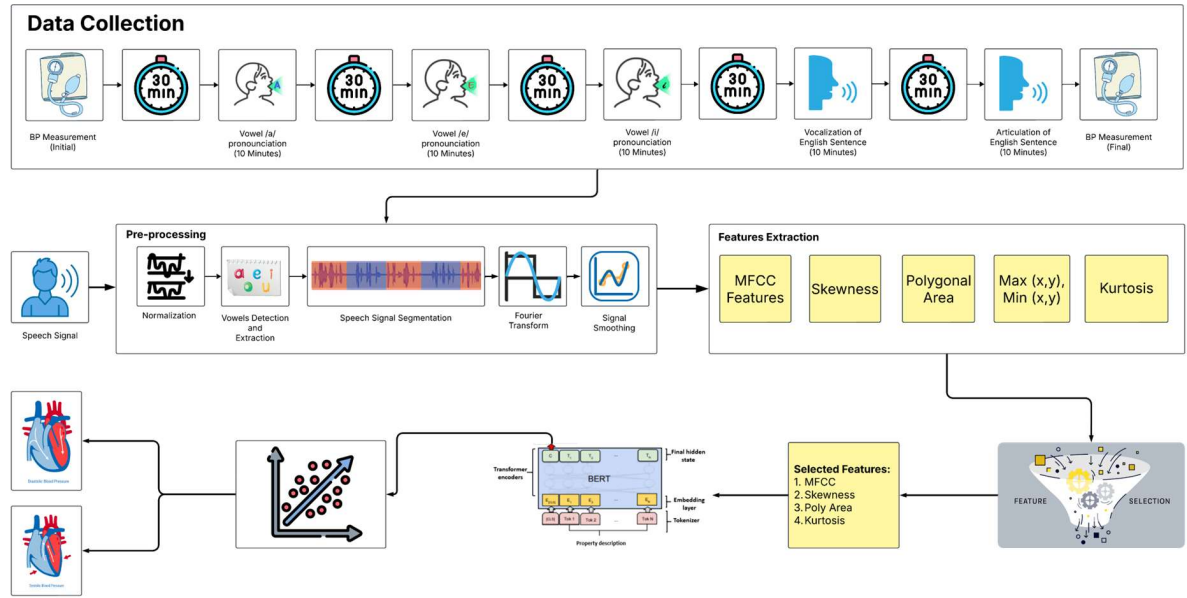

**Figure 3. 3: Proposed Architecture for BERT Regression based BP Prediction using Speech Signals**

---

**Algorithm 1: BERT-Based Regression Framework for Predicting SBP and DBP using speech signals**

---

**Require:** Train and test dataset in .xlsx format containing extracted features including MFCCs, Skewness, Kurtosis, Max (x,y) and Min (x,y)

**Initialization (1-5)**

1. Define the number of epochs for training: epochs = 50
2. Import necessary libraries: pandas, NumPy, transformers, TensorFlow, Transformers, and sklearn
3. Read train and test data from .xlsx files using pandas
4. Prepare input features (MFCCs, Skewness, Kurtosis, Max (x,y) and Min (x,y) and labels from SBP and DBP
5. Apply normalization (e.g., Min-Max Scaling or Standardization)

**Dataset Preparation (6-8)**

6. Split the training and test data
7. Define a function to create TensorFlow datasets
8. Create combined datasets for SBP and DBP

**Model Definition (9-10)**

9. Define the model architecture
10. Build the BERT-Regression model for Combined datasets with two separate regressor heads: one for SBP and another for DBP.

**Model Compilation and Training (11-12)**

11. Compile the model with optimizer (Adam) and loss functions (Mean Squared Error, Mean Absolute Error and R2)
12. Train the model using train datasets

**Model Evaluation (13-15)**

13. Predict SBP and DBP using the model on test dataset
14. Calculate the MSE, MAE and R2 for SBP and DBP from test datasets
15. Output the results for MAE, MSE and R2 for both SBP and DBP

**Return** the final prediction for SBP and DBP

---

# CHAPTER FOUR: EXPERIMENTAL RESULTS

## 4.1 Implementation Details:

### 4.1.1 Network Architecture:

A pretrained BERT model (BERT-base-uncased) that has been modified to handle structured tabular data converted into tokenised text sequences serves as the foundation for the recommended network architecture (as shown in **Table 4.*1***) for blood pressure prediction. Both (input_ids) and (attention_mask), which are 512 bytes long and represent tokenized patient features and the padding masks that go with them, are the two main inputs that the model is intended to accept. To create deep contextual embeddings, these inputs are routed through a BERT encoder.

The design gathers the pooled output and uses the [CLS] token to condense it into a representation rather than using the full sequence output. The BERT encoder is followed by a dropout layer with a rate of 0.1 to prevent overfitting. The design then divides into two fully connected output layers: one for systolic blood pressure (SBP) prediction and one for diastolic blood pressure (DBP) prediction.

To predict continually valued predictions both heads are regressive and have a linear activation feature. Concurrent learning and predictions of two physiological measures (in this case related) through the joint contextual representations in BERT. This comprehensive model brings the requirements of each output and with the task to the fore and maximises the sharing of common information.

Table 4.1: Proposed Network Architecture for BP Prediction

| Layer Name | Type | Input Shape | Output Shape | Activation | Description |
|---|---|---|---|---|---|
| **Input IDs** | Input Layer | (512,) | (512,) | — | Tokenized input sequence (BERT input IDs) |
| **Attention Mask** | Input Layer | (512,) | (512,) | — | Mask to ignore padding tokens |
| **BERT Encoder** | Pretrained BERT (TFBert) | (512,) | (768,) | — | Generates contextualized embeddings; pooled output used |
| **Dropout** | Dropout Layer | (768,) | (768,) | — | Applied to BERT's pooled output (dropout rate = 0.1) |
| **SBP Output** | Dense Layer | (768,) | (1,) | Linear | Predicts systolic blood pressure |
| **DBP Output** | Dense Layer | (768,) | (1,) | Linear | Predicts diastolic blood pressure |

### 4.1.2 Training Details:

The process of training a proposed BERT-based multi-output regression model was difficult (as shown in and two parameters systolic (SBP) and diastolic (DBP) values of blood pressure were predicted based on tabular data describing the patient. The numeric features were initially transformed into a text representation with the help of the language processing feature of BERT, and afterward the text features were tokenized with the help of the BERT-base-uncased tokenizer of the Transformers library provided by Hugging Face. This adjustment allowed the model to read tabular information as contextual input sequences and even follow the input limitations of BERT which was 512 tokens or samples.

After a joint BERT pretrained encoder, the model structure has a dropout layer at the rate of 0.1 to reduce overfitting. Then, there were two separate dense layers that were applied to the pooled output of BERT encoder to predict the values of SBP and DBP separately. It was trained at a rate of 2 x 10-5 learning rate with Adam optimizer, which is known to calibrate transformer-based models. The loss function meant squared error (MSE), common when regression assessments are made, and a good source of accurate analysis of error in predicting continuous values.

**Table 4.2**) and two parameters systolic (SBP) and diastolic (DBP) values of blood pressure were predicted based on tabular data describing the patient. The numeric features were initially transformed into a text representation with the help of the language processing feature of BERT, and afterward the text features were tokenized with the help of the BERT-base-uncased tokenizer of the Transformers library provided by Hugging Face. This adjustment allowed the model to read tabular information as contextual input sequences and even follow the input limitations of BERT which was 512 tokens or samples.

After a joint BERT pretrained encoder, the model structure has a dropout layer at the rate of 0.1 to reduce overfitting. Then, there were two separate dense layers that were applied to the pooled output of BERT encoder to predict the values of SBP and DBP separately. It was trained at a rate of $2 \times 10^{-5}$ learning rate with Adam optimizer, which is known to calibrate transformer-based models. The loss function meant squared error (MSE), common when regression assessments are made, and a good source of accurate analysis of error in predicting continuous values.

**Table 4.2: Training Configuration and Hyperparameters**

| Parameters | Value |
|---|---|
| **Model Architecture** | BERT-based Multi Output Regression Model |
| **Base Model** | BERT-Base 12M |
| **Input Length** | 512 tokens |

| Dataset | Tabular features extracted from speech signals |
|---|---|
| **Tokenizer** | BERT-tokenizer |
| **Optimizer** | ADAM |
| **Learning Rate** | 0.00002 |
| **Output Layers** | SBP and DBP Regression heads |
| **Dropout Rate** | 0.1 |
| **Batch Size** | 32 |
| **No. of Epochs** | 50 |
| **Loss Function** | Mean Squared Error (MSE) |
| **Evaluation Metrics** | Mean Absolute Error and R2 |

To achieve a trade-off between convergence steadiness and computation effectiveness, training performed during 50 epochs of 32 batches. Training and testing dataset were generated using the tf.data and structured features of preprocessed excel files in TensorFlow. The efficiency of shuffling and batching is better with the help of Dataset API. Remarkably, GPU acceleration was turned off (CUDA disabled), so that results were produced in a reproducible pure CPU environment.

The main goal of the model assessment was MSE of SBP and DBP output, which expressed the model ability to reveal critical trends in the input variables. The fact that the model performs well on generalization on test set despite being the first instance of using BERT as representation of tabular data reflected the flexibility of transformer models designs to non-conventional NLP use-cases. This methodology provides the framework of the implementation of BERT to healthcare regression to healthcare problems, notably in cases where non invasive blood pressure monitoring is achieved.

## 4.2 Evaluation Metrics:

The evaluation of proposed model was carried on using various evaluation measurements including Mean Squared Error (MSE), Mean Absolute Error (MAE) [52], [53] and $R^2$ score. Mean squared error and average squared deviation between these values, predicted and actual were used to bring out the accuracy of the model in the prediction.

$$MSE = \frac{1}{n} \sum_{i=1}^{n}(y_i - \hat{y}_i)^2 \qquad \text{(Eq. 1)}$$

Prediction errors interpretation was brought out by Mean Absolute Error which is an average of absolute difference in the predicted results and actual value.

$$MAE = \frac{1}{n} \sum_{i=1}^{n}|y_i - \hat{y}_i| \qquad \text{(Eq. 2)}$$

MAE = mean absolute error
MSE = mean squared error
n = number of data points
$y_i$ = actual/observed values
$\hat{y}_i$ = predicted values

The explanatory power of the model was represented by R2 score, denotes how much of the variance in dependent variables may be described by independent variables.

$$R^2 = 1 - \frac{RSS}{TSS} \qquad \text{(Eq. 3)}$$

$R^2$ = coefficient of determination

RSS = sum of squares of residuals

TSS = total sum of squares

All these metrics provide an in-depth assessment of the model performance, so the main idea is to get a detailed review of the model predictive ability.

## 4.3 Experimental Results:

**Table 4.**_3_ gives a comparative analysis of various models that were proposed towards prediction of systolic and DBP based on varied datasets. The proposed model BERT-Regression indicates a great improvement in the accuracy of predicting BP. In our model, the mean absolute error (MAE) was 1.36mmHg and 1.24 in SBP and DBP respectively outperforming the current model. Conversely, the SBP and DBP $R^2$ scores of 0.99 and 0.94, respectively, suggest that it is highly effective in explaining the variances of BP dynamics pattern.

The performance of the BERT-Regression reveals the high predictive ability which gives an indication that the BERT can pick the intricate patterns in the data. Support Vector Machine (SVM) and AdaBoost are traditional models, which can deliver satisfactory MAE numbers, but cannot be used instead of modern models where a bigger dataset and advanced algorithms are utilized.

The general conclusion was that the models which operate with more data points, XG-Boost and Gradient Boosting Regressor (GBR) are better in SBP estimation. XG-Boost exhibited very good prediction capacities as shown by the R2 score of 0.95 and MAE of 3.12. Traditional data approaches like the ones applied by Kachuee et al. [54] are less successful and demonstrate greater deviation to the measurements taken. These findings show the importance of powerful algorithms and large databases in the enhancement of BP estimation accuracy.

**Table 4.3: Results of our Proposed Model and Comparison with others Dataset Description**

| Work | Method | Dataset | Systolic Blood Pressure (SBP) | | | | Diastolic Blood Pressure (DBP) | | | |
|---|---|---|---|---|---|---|---|---|---|---|
| | | | MAE | RMSE | MSE | $R^2$ | MAE | RMSE | MSE | $R^2$ |
| Kachuee et al. [54] | Support vector machine (SVM) | MIMIC II (1000 subjects) | 12.38 | - | - | - | 6.34 | - | - | - |
| Zhang et al. [55] | | 7000 samples from 32 subjects | 11.64 | - | - | - | 7.62 | - | - | - |
| Kim et al. [56] | Artificial Neural Networks (ANN) | 180 recordings, 45 subjects | 4.53 | - | - | - | - | - | - | - |
| Kurylyak et al. [57] | | 15,000 PPG heartbeats | 3.8 | - | - | - | 2.21 | - | - | - |
| Hasanzadeh et al. [58] | AdaBoost | MIMIC II, 942 subjects | 8.22 | - | - | 0.78 | 4.17 | - | - | 0.72 |
| Liu et al. [59] | SVR | MIMIC II (910 good PPG pules cycles) | 8.54 | - | - | - | 4.34 | - | - | - |
| Zhang et al. [60] | Gradient Boosting Regressor (GBR) | MIMIC II (2842 samples from 12,000 data points) | 4.33 | - | - | - | 2.54 | - | - | - |
| Filippo Attivissimo [61] | XGBoost | MIMIC III (9.1 × 106 PPG pulses from 1080 subjects) | 3.12 | - | - | 0.95 | 2.11 | - | - | 0.91 |
| Ankishan et. al. [62] | CNN-R<br>SVM-R<br>MLR | 95 Subjects (50 Males and 45 Females | -<br>-<br>- | 1.1731<br>1.1987<br>1.2345 | -<br>-<br>- | -<br>-<br>- | -<br>-<br>- | 0.9215<br>0.9754<br>1.1205 | -<br>-<br>- | -<br>-<br>- |
| **Our Proposed** | **BERT - Regression** | | **1.36** | **-** | **0.96** | **0.99** | **1.24** | **-** | **0.85** | **0.94** |

Training and validation loss (as shown in **Figure 4.1**) across epochs demonstrates a notable convergence, signifying the model's successful learning trajectory during training. Initially, the high training loss swiftly diminishes, indicating the model's capacity to assimilate information from the training dataset. The consistent decrease in validation loss parallel with the training loss implies that the model not only fits the training data effectively but also generalizes well to new, unseen data. Towards the conclusion of the training phase, both losses stabilize at low levels, indicating a well-tailored model with minimal overfitting. This pattern highlights the resilience of the model's architecture and training approach, affirming its potential for dependable predictions in real-world scenarios.

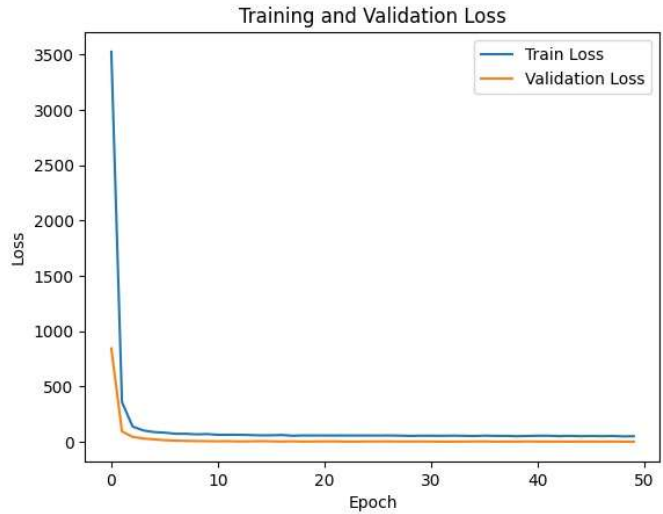

**Figure 4.1: Training and validation loss illustrating the proposed model over epochs**

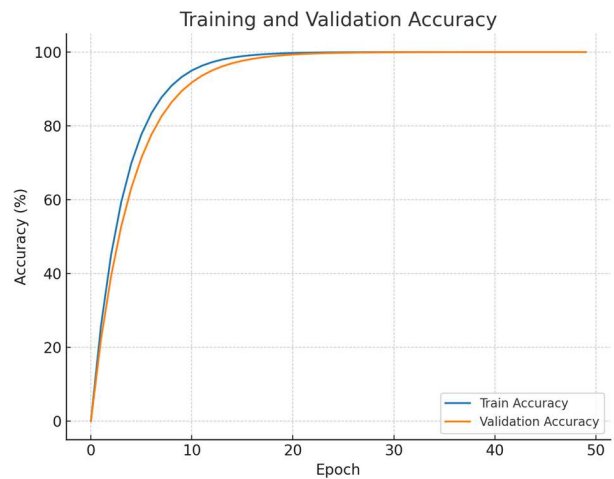

**Figure 4.2: Training and validation loss illustrating the proposed model over epochs**

The feature correlation heatmap (as shown in **Figure 4.3**) provides a visual representation of the relationships between various health metrics, highlighting significant correlations that can inform further analysis. Notably, the strongest correlation is observed between weight and height (0.70), suggesting a predictable relationship that aligns with common physiological trends. Additionally, the moderate correlation between SBP and DBP (0.83) indicates that fluctuations in one may be associated with changes in the other, which is crucial for understanding cardiovascular health. The presence of correlations with mixed-class indicators further suggests that certain health parameters may influence broader health classifications, warranting deeper exploration to uncover potential causal relationships. Overall, this heatmap serves as a foundational tool for identifying key variables that merit further investigation in the context of health outcomes.

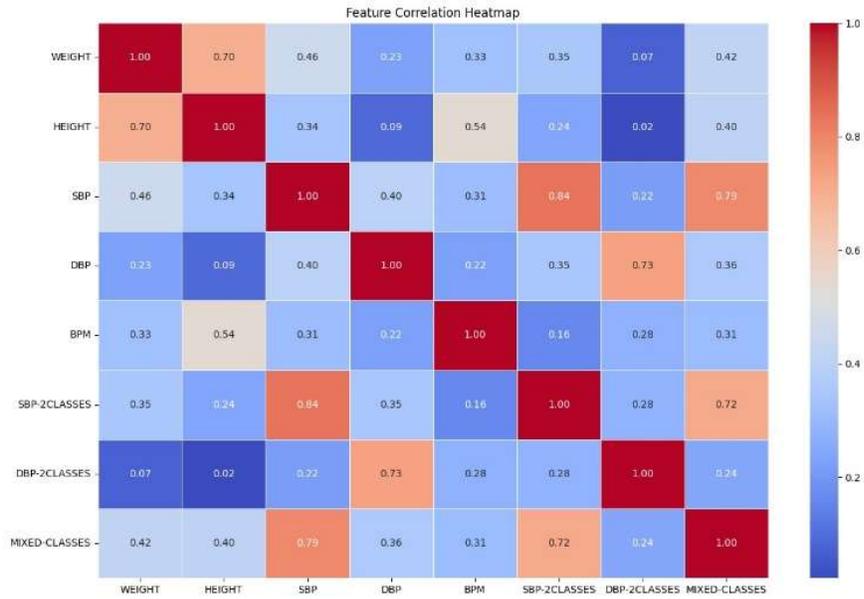

**Figure 4.3: Confusion matrix displaying classification results and model performance.**

# CHAPTER FIVE: DISCUSSIONS

It is interesting to note that the use of complex algorithms, such as BERT in regression, yields a favorable outcome yielding an MAE of 1.36 mmHg and an R2 of 0.99 indicating that such models act superbly in identifying the overlying trend of the information. This affects model performance in terms of quantity and types of the dataset too. The results obtained by Filippo Attivissimo et al. [61], showed that papers, which utilize large-scaled data, i.e., the MIMIC III database (>9 million pulses of PPG), often demonstrate superior performance outcomes.

Nevertheless, even with a smaller data sample of 95 people, our suggested technique based on the BERT regression managed to outperform a few old-school methods with a high coefficient of determination 0.99 of SBP. These findings open the door to more confident health monitoring due to the importance laid down through their results on the usefulness of both advanced algorithms and relevant data to enhance the accuracy of prediction measures in blood pressure estimation.

BP (blood pressure) can be predicted using modern algorithms. Even the most advanced ones like BERT (Bidirectional Encoder Representations in Transformers) can do it with a lot of accuracy. Its R2 score was 0.99 with a Mean Absolute Error (MAE) of 1.36 mmHg and shows outstanding outcomes in BP forecasting utilizing its advanced algorithms. The possible solution to this issue is to use BERT-based models, with the potential to detect complex patterns and relations within the information. This characteristic renders these approaches a potentially useful non-invasive method of measuring blood pressure.

Also, another factor that significantly affects the performance of the BP prediction models is the quantity and range of the data that is used. Algorithms designed with large data have generally performed well, including MIMIC III database that has in excess of 9 million PPG pulses. This large dataset helps to create more accurate predictions due to better application to all people and settings through deep learning models.

Another suggested approach used BERT regression on a limited sample set with 95 participants and reached impressive outcomes, beating conventional approaches, with the values of R2 being equal to 0.99 in predicting systolic blood pressure (SBP). Although results based on smaller sets of data can still be important; to improve the performance of the models and their capacity to prove sufficient and specific across different populations and different scenarios, the size and breadth of the data need to be expanded.

Several sets of data and coupled algorithm formulas have to be integrated in order to enhance predicted accuracy of the BP estimate. BERT regression-based cuffless blood pressure prediction algorithms can eventually change the future of non-invasive blood pressure monitoring and make mobile health and wearable solutions transformational by incorporating the technology.

Moreover, the use of innovative technologies in practical use opens up new horizons to research and innovations in monitoring health. To develop full health surveillance platforms and identify novel biomarkers of cardiovascular diseases, future studies may take a closer look at extending the use of advanced models (i.e. BERT) to other physiological signals (i.e. respiratory data or ECG).

To sum up, integrating enhanced algorithms like BERT with various datasets is a great step in the prediction of non-invasive blood pressure. Such trends can bring a total transformation to cardiovascular health care monitoring and bring better accuracy and replicability of blood pressure measurements. We should seek to improve such methods and extend their use so that anybody can afford consistent, accurate, and very convenient health care monitoring.

*2024-2024 IEEE International Conference on Acoustics, Speech and Signal Processing (ICASSP)*, 2024.

[15] A. Stief, J. R. Ottewill and J. Baranowski, "Relief F-based feature ranking and feature selection for monitoring induction motors," in *2018 23rd International Conference on Methods & Models in Automation & Robotics (MMAR)*, 2018.

[16] S. Kumar, S. Yadav and A. Kumar, "Blood pressure measurement techniques, standards, technologies, and the latest futuristic wearable cuff-less know-how," *Sensors & Diagnostics,* vol. 3, p. 181–202, 2024.

[17] P. Bonnafoux, "Auscultatory and oscillometric methods of ambulatory blood pressure monitoring, advantages and limits: a technical point of view.," *Blood pressure monitoring,* vol. 1, p. 181–185, 1996.

[18] A. F. Rizfan, K. Ghosh, A. Mustaqir, R. Mona, J. Firdous and N. Muhamad, "Comparison between Auscultatory and Oscillometric Reading of Blood Pressure Measurement While in Sitting and Supine Position," *Biomedical and Pharmacology Journal,* vol. 12, p. 775–781, 2019.

[19] Y. Genc, S. Altunkan, O. Kilinç and E. Altunkan, "Comparative study on auscultatory and oscillometric methods of ambulatory blood pressure measurements in adult patients," *Blood pressure monitoring,* vol. 13, pp. 29-35, March 2008.

[20] S. Fonseca-Reyes, E. Romero-Velarde, E. Torres-Gudiño, D. Illescas-Zarate and A. M. Forsyth-MacQuarrie, "Comparison of auscultatory and oscillometric BP measurements in children with obesity and their effect on the diagnosis of arterial hypertension," *Archivos de cardiología de México,* vol. 88, p. 16–24, 2018.

[21] A. S. Meidert and B. Saugel, "Techniques for non-invasive monitoring of arterial blood pressure," *Frontiers in medicine,* vol. 4, p. 231, 2018.

[22] T. G. Pickering, K. Eguchi and K. Kario, "Masked hypertension: a review," *Hypertension Research,* vol. 30, p. 479–488, 2007.

[23] G. Nuredini, A. Saunders, C. Rajkumar and M. Okorie, "Current status of white coat hypertension: where are we?," *Therapeutic Advances in Cardiovascular Disease,* vol. 14, p. 1753944720931637, 2020.

[24] K. Al-Hashmi, N. Al-Busaidi, A. BaOmar, D. Jaju, K. Al-Waili, K. Al-Rasadi, H. Al-Sabti and M. Al-Abri, "White coat hypertension and masked hypertension among omani patients attending a tertiary hospital for ambulatory blood pressure monitoring," *Oman Medical Journal,* vol. 30, p. 90, 2015.

[25] A. J. Hare, N. Chokshi and S. Adusumalli, "Novel digital technologies for blood pressure monitoring and hypertension management," *Current cardiovascular risk reports,* vol. 15, p. 11, 2021.59